\pdfoutput=1

\documentclass[11pt]{article}

\usepackage[]{EMNLP2023}

\usepackage{times}
\usepackage{latexsym}

\usepackage[T1]{fontenc}

\usepackage[utf8]{inputenc}

\usepackage{microtype}

\usepackage{inconsolata}

\usepackage{multirow}
\usepackage{booktabs}
\usepackage{graphicx}
\usepackage{subfigure}
\usepackage{amssymb}
\usepackage{CJKutf8}
\usepackage{amsmath} 
\usepackage{hyperref}
\usepackage{enumitem}
\usepackage{color}
\graphicspath{{images/}}
\usepackage{algorithm}
\usepackage{algpseudocode}

\title{What Causes the Failure of Explicit to Implicit Discourse \\ Relation Recognition?}

\author{Wei Liu$^{1}$, Stephen Wan$^{2}$, Michael Strube$^{1}$ \\ $^{1}$Heidelberg Institute for Theoretical Studies gGmbH  \\ $^{2}$CSIRO Data61 \\ \texttt{wei.liu@h-its.org} | \texttt{stephen.wan@csiro.au} | \texttt{michael.strube@h-its.org}}

\begin{document}
\maketitle

\begin{abstract}
We consider an unanswered question in the discourse processing community: why do relation classifiers trained on explicit examples (with connectives removed) perform poorly in real implicit scenarios? Prior work claimed this is due to linguistic dissimilarity between explicit and implicit examples but provided no empirical evidence. In this study, we show that one cause for such failure is a label shift after connectives are eliminated. Specifically, we find that the discourse relations expressed by some explicit instances will change when connectives disappear. Unlike previous work manually analyzing a few examples, we present empirical evidence at the corpus level to prove the exist\-ence of such shift. Then, we analyze why label shift occurs by considering factors such as the syntactic role played by connectives, ambiguity of connectives, and more. Finally, we investigate two strategies to mitigate the label shift: filtering out noisy data and joint learning with connectives. Experiments on PDTB 2.0, PDTB 3.0, and the GUM dataset demonstrate that classifiers trained with our strategies outperform strong baselines.
\end{abstract}

\section{Introduction}
Discourse relations, such as \textit{Contrast} and \textit{Cause}, describe the logical relationship between two text spans (i.e., arguments). They can either be signaled explicitly with connectives, as in (1), or expressed\- implicitly, as in (2):

\vspace{0pt}
\begin{figure}[h]
\centering\includegraphics[scale=0.49,trim=0 0 0 0]{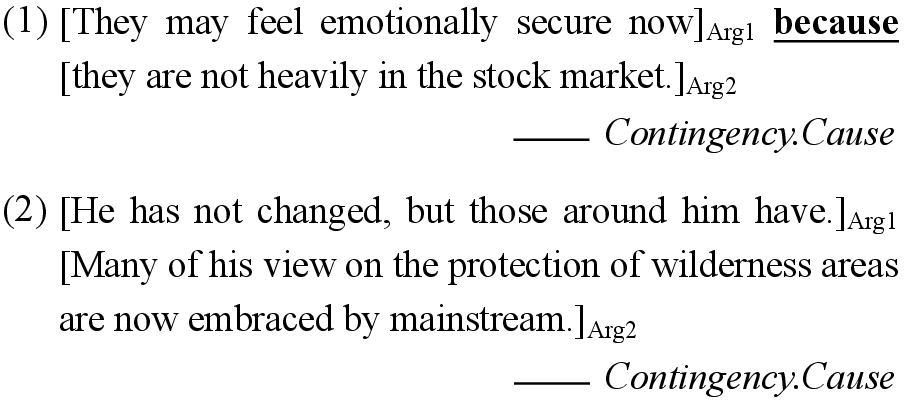}
\vspace{-18pt}
\end{figure}

\noindent Corpora of explicit discourse relations are relatively easy to create manually and automatically~\citep{marcu-echihabi-2002-unsupervised} since connectives are strong cues for identifying relations~\citep{pitler-nenkova-2009-using}. In contrast, the annotation of implicit relations is hard and expensive because one must infer the sense from input texts. This has prompted many early studies~\citep{marcu-echihabi-2002-unsupervised, lapata-lascarides-2004-inferring, sporleder_2005, saito-etal-2006-using} to use explicit examples to classify implicit relations (dubbed \textbf{explicit to implicit relation recognition}). The main idea is to construct an \textit{implicit-like} corpus by removing connectives from explicit instances, and use it to train a classifier for implicit relation recognition. While this method attains good results on test sets constructed in the same manner, it is reported by~\citet{sporleder_lascarides_2008} to perform poorly in real implicit scenarios. They claim this is caused by the linguistic dissimilarities between explicit and implicit examples, but provide no corpus-level empirical evidence. More recent works~\citep{huang-li-2019-unsupervised, kurfali-ostling-2021-lets} focus on enhancing the transfer performance from explicit to implicit discourse relations. However, little attention has been paid to the underlying causes of the poor results.

In this paper, we show that one cause for the poor transfer performance is the occurrence of label shift during the construction of the \textit{implicit-like} corpus. Removing connectives from explicit examples affects the discourse relations they originally expressed. 
Referring to example (3), which contains the connective \textit{then} and is annotated as a \textit{Temporal.Asynchronous} relation:

\vspace{-2pt}
\begin{figure}[h]
\centering\includegraphics[scale=0.50,trim=0 0 0 0]{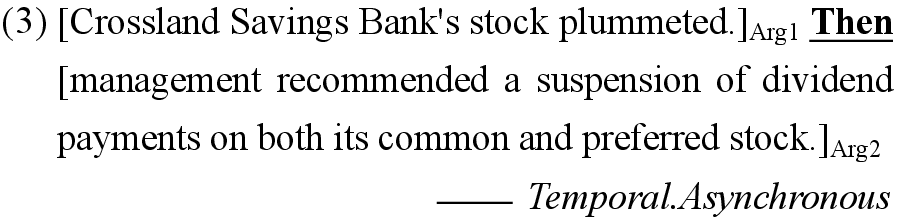}
\vspace{-22pt}
\label{fig:intro-exam}
\end{figure}

\noindent When the connective \textit{then} is removed, the example, however, tends to express a \textit{Contingency.Cause} relation because the first argument describes a result of "stock plummet" and the second argument gives the reason, a suspension of dividend pay. To verify the existence of the label shift, we first manually analyze a small number of explicit examples with connectives removed, and summarize different cases of instances suffering such shift. Then, we provide empirical evidence to demonstrate that label shift is present not only in a few examples but at the corpus level. Consequently, classifiers trained on the \textit{implicit-like} corpus learn a chaotic pattern for relation classification (i.e., being taught to predict an example with \textit{Cause} relation as \textit{Asynchronous} relation as in the above example), resulting in poor performance in real implicit scenarios. We further analyze why label shift happens in the \textit{implicit-like} corpus by considering factors such as the syntactic role played by connectives, ambiguity of connectives, and more. Our results reveal that the syntactic role played by connectives contributes the most to the occurrence of the label shift.

Based on this observation, we investigate two strategies from both the data and training aspects to alleviate the influence of the label shift. We devise a label shift metric to quantify the degree of label shift that occurs in each explicit example and employ it for sample-level filtering. Additionally, we study a joint-learning strategy from the training side to further alleviate the impact of the shift in cases where the filtering results are imperfect. Specifically, our classifier jointly learns to recover a connective from arguments and identify a relation based on the recovered connective and arguments.
 
We evaluate the effectiveness of our approach on two datasets: the Penn Discourse Treebank 2.0~\citep[PDTB 2.0, ][]{prasad-etal-2008-penn} and 3.0~\citep[PDTB 3.0,][]{webber2019penn}. Experiments show that our model improves the performance of explicit to implicit discourse relation recognition, achieving encouraging results on both datasets. Furthermore, to test the generalizability of the proposed method, we conduct experiments on the GUM dataset~\citep{gum-rst}, which is annotated with relations from Rhetorical Structure Theory~\citep[RST,][]{RST}. The results suggest that our filtering mechanism and joint training strategy also help with the explicit to implicit relation classification on the GUM dataset.

\section{Related Work}
Learning to use lexically-marked examples to classify implicit relations has received continued research attention.~\citet{marcu-echihabi-2002-unsupervised} train the first classifier for implicit intra-sentential discourse relations using explicitly-marked examples from a raw English corpus, BLIPP~\citep{charniak-2000-maximum}, and the RST Treebank~\citep{carlson-etal-2001-building}.~\citet{lapata-lascarides-2004-inferring} present a similar approach using BLIPP but focus on sentence-internal temporal relations.~\citet{blair-goldensohn-etal-2007-building} extend this work by refining the training process using parameter optimization, topic segmentation, and syntactic parsing on the Gigaword~\citep{graff2003english} and PDTB~\citep{pdtb1}. These three works are evaluated on test sets constructed in the same manner as the training set and show good performance.~\citet{sporleder_lascarides_2008} and~\citet{lin-etal-2009-recognizing} investigate the applicability of this approach to real implicit scenarios and find its performance is poor. They claim, based on manual analysis of a few instances, that the linguistic dissimilarities between explicit and implicit examples may be the cause. However, a corpus-level empirical analysis is not provided to establish how widespread the problem is.

More recent work has focused on improving the performance from explicit to implicit discourse relation recognition.~\citet{wang-etal-2012-implicit} propose to use typical examples with linguistic structure shared between explicit and implicit relations for training.~\citet{ji-etal-2015-closing} adopt techniques such as resampling from transfer learning to handle the mismatched label distribution between explicit and implicit corpora.~\citet{huang-li-2019-unsupervised} follow a similar domain adaptation idea but focus on minimizing the distance between representations of explicit and implicit examples with an adversarial training framework.~\citet{kurfali-ostling-2021-lets} tackle this task from a distant-supervision perspective. In contrast to the above work, we aim to present a new understanding of the question "Why classifiers trained on explicit examples perform poorly in real implicit scenarios" and to provide corpus-level empirical evidence to support our findings.

\begin{figure*}[t]
\centering\includegraphics[scale=0.383,trim=0 0 0 0]{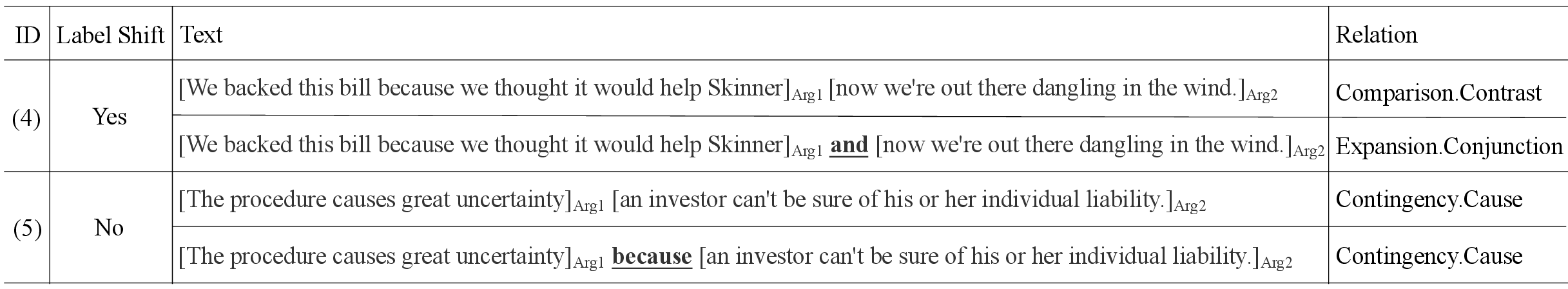}
\setlength{\abovecaptionskip}{10pt}
\setlength{\belowcaptionskip}{-6pt}
\caption{Examples of suffering and not suffering the label shift.}
\label{fig:sens-exam}
\end{figure*}

Connective information has been widely studied in discourse relation recognition.~\citet{pitler-nenkova-2009-using} train a classifier with connectives in the text as the only features and find it could achieve over 90\% accuracy on explicit relation recognition. Similarly, many attempts have been made using connectives to improve the recognition performance on implicit relations, including pipeline methods~\citep{zhou-etal-2010-predicting,pipejiang}, multi-task training~\citep{kishimoto-etal-2020-adapting, long-webber-2022-facilitating}, adversarial training~\citep{qin-etal-2017-adversarial}, joint training~\citep{liu-strube-2023-annotation}, and prompt learning~\citep{zhou-etal-2022-prompt-based,xiang-etal-2023-teprompt}. Our work differs from them in both motivation and application scenarios. For example, we use connective information not as a feature to the classifier but as a filtering mechanism to select good training instances.

\section{Experimental Setup}
\label{sec:exp-set}
We introduce the task of explicit to implicit relation recognition and describe the experimental setup\footnote{\url{https://github.com/liuwei1206/Exp2Imp}} used for analyses in Section \ref{sec:label_shift} and improvements in Section \ref{sec:exp_res}.

\noindent \textbf{Task.} The task of explicit to implicit relation recognition builds an implicit classifier relying on explicit examples. The traditional approach to achieving this is to construct an \textit{implicit-like} corpus by excluding connectives from explicit examples, and then train a classifier on this corpus with the original explicit relations as ground-truth labels.

\noindent \textbf{Datasets.}
The datasets we use for analyses are PDTB 2.0~\citep{prasad-etal-2008-penn} and PDTB 3.0~\citep{webber2019penn}. PDTBs are corpora annotated with a lexical-based framework where instances are divided into different groups, including the discourse relation categories we focus on: explicit and implicit. This clear grouping makes them very suitable for explicit to implicit relation recognition~\citep{huang-li-2019-unsupervised, kurfali-ostling-2021-lets} since we do not need to distinguish explicit or implicit examples by ourselves. More import\-antly, the two corpora offer manually annotated connectives for implicit examples (See Appendix \ref{app:dataset}), facilitating our comparative analysis of explicit and implicit\- relations.

We follow previous work~\citep{huang-li-2019-unsupervised} to use PDTB sections 2-20, 0-1, and 21-22 as training, development, and test set, respectively. We conduct experiments on both the top- and second-level relations of the two corpora.

\noindent \textbf{Models.}
The relation classifier employed in this paper, including models for analysis in section \ref{sec:label_shift} and baselines in section \ref{sec:exp_res}, consists of a pre-trained encoder and a linear layer. We follow previous work~\citep{zhou-etal-2022-prompt-based,long-webber-2022-facilitating} to use $\rm RoBERTa_{base}$ as the encoder. We show in Appendix \ref{app:diff_type_size} that our findings are consistent across different pre-trained models and sizes. See Appendix \ref{app:setup} for more detailed settings.

\section{Label Shift in Discourse Relations}
\label{sec:label_shift}

\subsection{What is label shift?}
We consider label shift as the difference in relations observed between the same example with and without a connective:
\begin{equation}\label{eqt:shift}
    {\rm Rel(Arg1, Conn, Arg2) \neq Rel(Arg1, Arg2)}
\end{equation}
where $\rm Arg1$ and $\rm Arg2$ are arguments of the example, and $\rm Conn$ denotes the connective. Figure \ref{fig:sens-exam} shows examples of suffering and not suffering the label shift. Example (4) was originally annotated as an \textit{Expansion.Conjunction} relation because of the connective \textit{and}. When \textit{and} is removed, the example tends to express a \textit{Comparison.Contrast} relation because of the contrast in lexical cues (e.g., "would help" vs. "dangling in the wind"). Regarding example (5), the arguments express the same relation as the connective \textit{because} since the first argument describes a result of "uncertain" and the second argument presents the reason, i.e., unsure of liability.

\subsection{Do explicit examples suffer label shift?}
\label{sec:suffer-shift}

We manually analyze 100 explicit instances in PDTB 2.0 to ascertain the existence of label shift. Specifically, we randomly sample 100 explicit examples from the PDTB 2.0 and remove the connectives. Then, we train two students to annotate raw texts with PDTB relations. After finishing the training, the two students are asked to annotate those 100 examples (with connective removed), and the inter-annotator agreement is 0.7346 calculated in Cohen's Kappa.\footnote{We use $\rm cohen\_kappa\_score$ in sklearn.} See Appendix \ref{app:manu_ana} for more annotation information. 

\begin{figure}[t]
\centering\includegraphics[scale=0.49,trim=0 0 0 0]{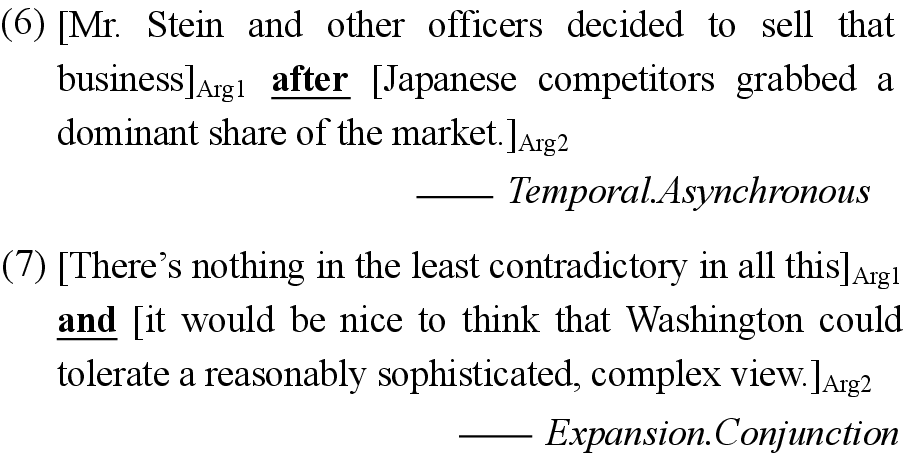}
\setlength{\abovecaptionskip}{-4pt}
\setlength{\belowcaptionskip}{-8pt}
\caption{Different cases suffering label shift.}
\label{fig:shift-exam}
\end{figure}

We find that 37 of these 100 examples were annotated with relations different from the original annotation, suggesting the existence of the label shift. We identify three different cases of suffering label shift: (i) Removing connectives leads to different relations. Referring to example (3), where the connective \textit{then} signals a \textit{Temporal} relation while the arguments express a \textit{Contingency} relation because the first argument describes a result of "stock plummet" and the second one points out the reason, a suspension of dividend pay. (ii) Deleting connectives causes ambiguity in relations. This occurs when arguments contain clues to multiple relations without favoring a certain one. Considering example (6) in Figure \ref{fig:shift-exam}, the arguments can express \textit{Contingency} or \textit{Temporal} relations since inserting \textit{because} or \textit{after} between arguments is acceptable. (iii) No relation is observed after eliminating the connective. This happens when there are no clues indicating discourse relations or arguments are too short to provide sufficient context. Referring to examples (7) in Figure \ref{fig:shift-exam}, there is low lexical cohesion between the two arguments, requiring extensive world knowledge to understand that "Washington" refers to the U.S. government and "politics" can be "complex" or "contradictory", making it hard to infer any relation. 

\subsection{Does this shift exist at the corpus level?}
\label{sec:ls_corpus}
We devise an empirical approach to show that label shift exists at the corpus level. The key idea comes from our definition of label shift, where an example is considered as suffering shift if its expressed relations are different when containing or not containing a connective. We mimic this judgment process but replace relations inferred by humans with those predicted by relation classifiers. Specifically, given a corpus with connectives (either explicit corpus or implicit corpus with implicit connectives), we first train a relation classifier using arguments-label pairs\footnote{We did not use examples with connectives to train classifiers because models trained in this way rely heavily on connectives for prediction~\citep{pitler-nenkova-2009-using}. By contrast, classifiers trained on arguments (no connectives) make predictions grounded in the semantics of examples.} of the corpus. Then, we compare the classifiers' predictions on this corpus with and without the use of connectives (i.e., explicit examples vs. explicit examples with connectives removed, or implicit examples with connectives vs. implicit examples). If the predictions in the two settings are very different (see $\rm diff\_num$ in Algorithm \ref{al:measuring-ls}), it implies that connectives can substantially affect the semantics of examples throughout the corpus. That is, label shift exists across the entire dataset.

\setlength{\textfloatsep}{8pt}
\begin{algorithm}[t]
    \small
    \renewcommand{\algorithmicrequire}{\textbf{Input:}}
    \renewcommand{\algorithmicensure}{\textbf{Output:}}
    \caption{Measuring Label Shift}
    \begin{algorithmic}[1]
        \Require
            Relation Classifier $\rm \mathbf{M}$, Corpus with Connectives $\rm \{(Arg1_i, Conn_i, Arg2_i, Rel_i)\}|_{i=1}^N$ 
        \Ensure
            $\rm diff\_num, scores$
            \State $\rm Train(\textbf{M}, \{(Arg1_i, Arg2_i, Rel_i)\}|_{i=1}^N)$
            \State $\rm diff\_num=0$ 
            \State $\rm scores=[]$ 
            \For{$\rm i=1, \dots, N$}
                \State \# without and with connectives
                \State $\rm p_1\, = \mathbf{M}.pred(Arg1_i, Arg2_i)$
                \State $\rm p_2\, = \mathbf{M}.pred(Arg1_i, Conn_i, Arg2_i)$
                \State $\rm \mathbf{v1} = \mathbf{M}.get\_rep(Arg1_i, Arg2_i)$                 
                \State $\rm \mathbf{v2} = \mathbf{M}.get\_rep(Arg1_i, Conn_i, Arg2_i)$ 
                \If{$\rm p_1 \neq p_2 $}
            \State $\rm diff\_num = diff\_num+1$
                \EndIf
                \State $\rm \,value  = cosine\_similarity(\mathbf{v1}, \mathbf{v2})$
                \State $\rm Append(scores, value)$
            \EndFor
    \end{algorithmic}
    \label{al:measuring-ls}
\end{algorithm}

\begin{figure}[t]
\vspace{8pt}
\centering\includegraphics[scale=0.47,trim=0 0 0 0]{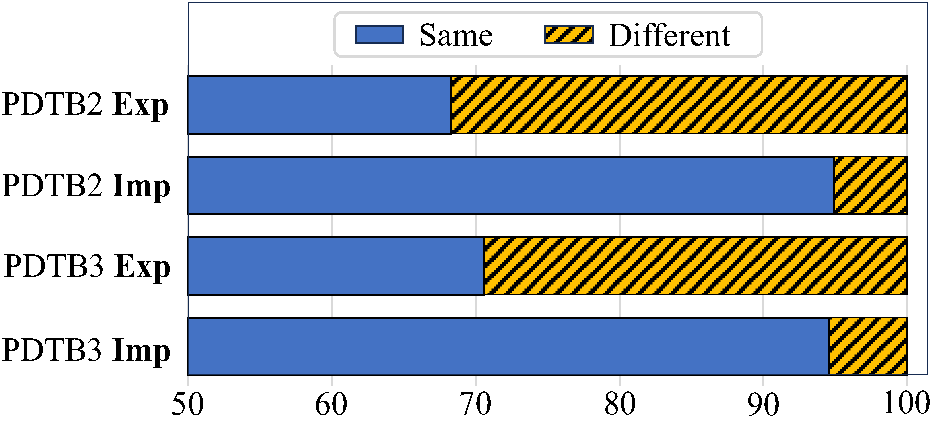}
\setlength{\abovecaptionskip}{8pt}
\setlength{\belowcaptionskip}{6pt}
\caption{Percentage of examples in \textbf{Exp}licit and \textbf{Imp}\-licit corpora that receive the same and different predictions when containing and not containing a connective.}
\label{fig:sen_non-sen}
\end{figure}

We conduct analyses on both explicit and implicit parts of PDTB 2.0 and 3.0, providing\- a comparison between these two types of examples. Figure\- \ref{fig:sen_non-sen} shows the assessment results on PDTB 2.0 and 3.0 (on top-level relations). In explicit corpora, connectives are more likely to influence the predictions of relation classifiers, with approximately 30\% of the examples being predicted as different relations when containing and not containing a connective. By contrast, only about 5\% of instances in the implicit corpora are predicted in different relations under the same settings. 

\begin{figure}[t]
\centering\includegraphics[scale=0.39,trim=0 0 0 0]{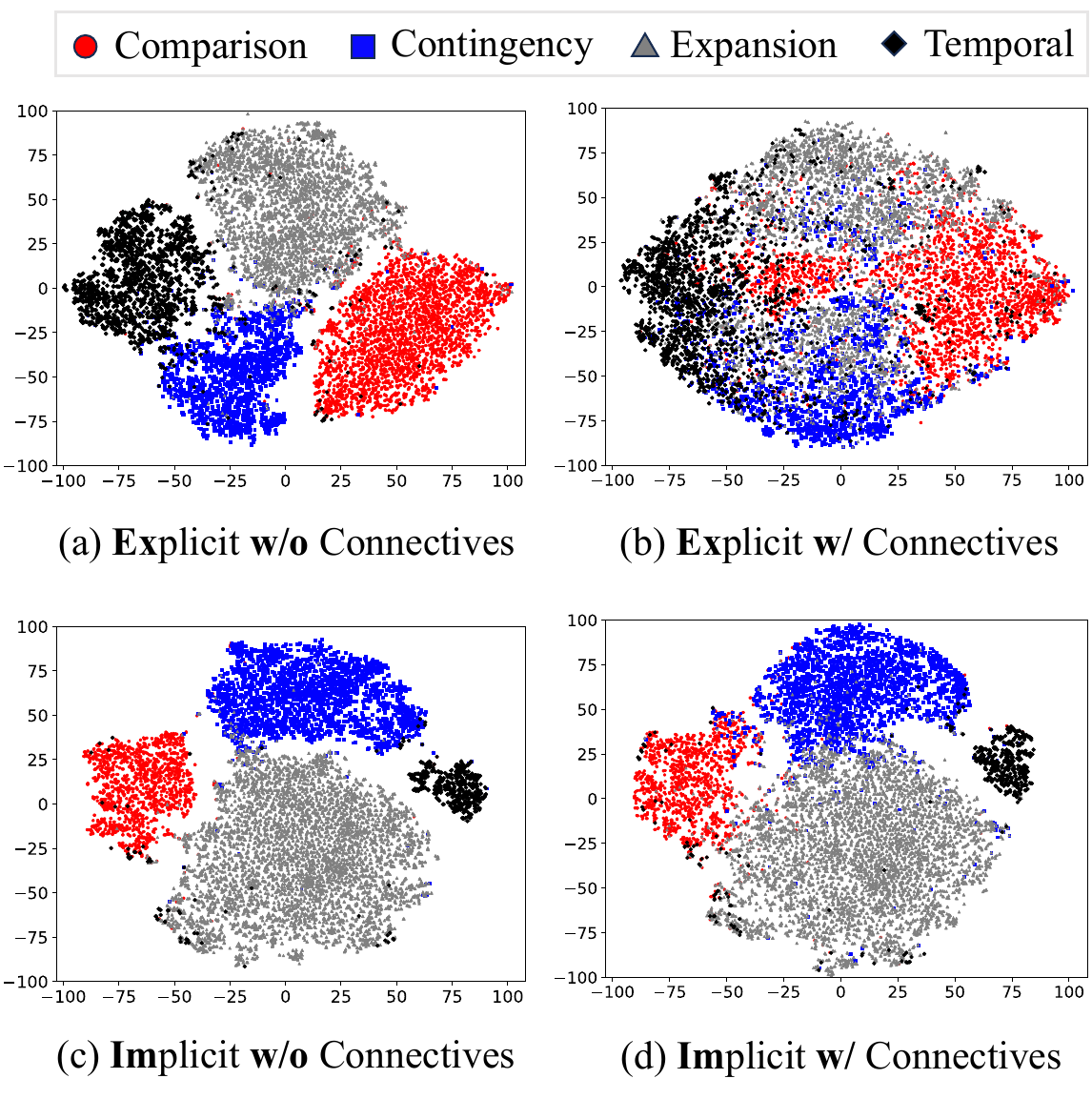}
\setlength{\abovecaptionskip}{-6pt}
\setlength{\belowcaptionskip}{6pt}
\caption{Visualization of examples in PDTB 2.0 when containing or not containing a connective.}
\label{fig:tnse-label-shift}
\end{figure}

We further visualize the representations of examples with and without a connective (see $\rm \mathbf{v1}$ and $\rm \mathbf{v2}$ in Algorithm \ref{al:measuring-ls}). Figure \ref{fig:tnse-label-shift} shows the visualized results on the training set of PDTB 2.0 (top-level relation) using t-SNE~\citep{vanDerMaaten2008}. Without connectives (see Fig \ref{fig:tnse-label-shift}a), explicit examples are well separated since the classifier is trained on arguments-label pairs. When inserting explicit connectives into inputs (see Fig \ref{fig:tnse-label-shift}b), the representations undergo significant changes, intertwining examples of different relations. Compared to the explicit cases, the representations of implicit instances generally remain unchanged after incorporating connectives (see Fig \ref{fig:tnse-label-shift}c and \ref{fig:tnse-label-shift}d), suggesting relations expressed by implicit arguments are barely affected by connectives.

The above results indicate that, after removing connectives, many examples in the explicit corpus express relations that differ from the original annotation. Consequently, classifiers trained on explicit examples (with connectives removed) learn a chaotic pattern for relation prediction, resulting in poor performance in real implicit scenarios.

\subsection{Can label shift be measured?}
\label{sec:measure-ls}
Different explicit instances exhibit varying degrees of label shift. For example, the case (i) in Section \ref{sec:suffer-shift} is more severe than case (ii) as deleting the connective makes the former convey a completely different relation (\textit{Temporal} $\to$ \textit{Contingency}) while rendering the latter ambiguous (but the original relation holds).  We design a \textbf{label shift metric} to quantify the degree of label shift that occurs in each instance of an explicit corpus. We show in Sections \ref{sec:why_label} and \ref{sec:filter_out} that this metric can be used to analyze factors causing label shift and to filter out noisy examples that suffer label shift, respectively. 

Given an explicit corpus with annotated relations $\rm \{(Arg1_i, Conn_i, Arg2_i, Rel_i)\}|_{i=1}^{N}$, we first train a classifier with arguments-relation pairs. Then, for each example, we extract representations of that example when containing and not containing a connective from the trained classifier's encoder, and calculate the cosine similarity between these two representations (see $\rm value$ in Algorithm \ref{al:measuring-ls}). If the cosine similarity is close to 1, it indicates that the example with and without connectives are semantically similar, thus suggesting the connective is more likely removable; otherwise, removing the connective probably results in a label shift. We compute the label shift metric for explicit corpora of PDTB 2.0 and 3.0, and find that around 33\% of explicit examples in PDTB 2.0 and about 29.6\% of those in PDTB 3.0 have a cosine similarity of less than 0.5, suggesting a substantial portion of connectives in the explicit dataset are not removable.

\subsection{Why does label shift happen?}
\label{sec:why_label}

While we have demonstrated that label shift occurs during the construction of the \textit{implicit-like} corpus, we know little about why removing a connective has such a significant impact. We investigate four factors that can contribute to label shift: (i) Is the removed connective a conjunction or an adverb~\citep{Prasad2006ThePD}? Conjunctions join clauses of equal grammatical rank in a sentence or join a subordinate clause to a main clause~\citep{sub-and-coor}. Removing conjunctions disrupts the syntactic structure of the texts and may make the relations expressed unclear. (ii) Is the removed connective ambiguous~\citep{webber-etal-2019-ambiguity}? Some connectives, such as \textit{since}, are ambiguous and signal multiply relations, which may result in the annotated relations of explicit examples being different from relations inferred from the arguments of these examples. (iii) Is the status of the arguments intra- or inter-sentential~\citep{prasad-etal-2018-discourse}? The information carried by intra-sentential arguments is incomplete (only parts of a sentence) and may not indicate a clear relation without the help of connectives. (iv) What is the length of the input arguments? Sufficient information is the key to inferring relations from text. If the arguments are very short, it will be hard to infer a relation in the absence of connectives. We extract these four features for each example in the explicit corpus, where the first three are represented as Boolean values (i.e., 0 or 1), and the last is represented as a floating value (normalize the length to a value between 0 and 1).

\begin{table}[t]
\centering
\scalebox{0.78}{
\begin{tabular}{l|cccc}
\hline
\multirow{2}{*}{} & \multicolumn{2}{c|}{PDTB 2.0}                        & \multicolumn{2}{c}{PDTB 3.0}   \\ \cline{2-5} 
                  & coefficient & \multicolumn{1}{c|}{p-value}          & coefficient & p-value          \\ \hline
Conj vs. Adv & -0.3946       & \multicolumn{1}{c|}{\textless{}0.001} & -0.3226       & \textless{}0.001 \\
Ambiguity & -0.0981       & \multicolumn{1}{c|}{\textless{}0.001} & -0.0412       & \textless{}0.001 \\
Intra vs. Inter & -0.1947      & \multicolumn{1}{c|}{\textless{}0.001} & -0.1898       & \textless{}0.001 \\
Input length      & \hspace{0.35em}0.1416      & \multicolumn{1}{c|}{\textless{}0.001} & \hspace{0.35em}0.1944      & \textless{}0.001 \\ \hline
\end{tabular}}
\setlength{\abovecaptionskip}{10pt}
\setlength{\belowcaptionskip}{4pt}
\caption{Pearson correlation between each individual factor and the label shift metric.}
\label{table:correlation}
\end{table}

We calculate the Pearson correlation between each individual factor and the label shift metric calculated in Section \ref{sec:measure-ls}, and show the results on PDTB 2.0 and 3.0 (top-level relation) in Table \ref{table:correlation}. All factors are significantly correlated with the label shift metric (p-value < 0.001) but with different correlation coefficients. The syntactic role played by connectives receives the largest value (in terms of absolute value), indicating whether the removed connective is a conjunction or an adverb has the most impact on the occurrence of label shift. It is followed by the status and length of arguments. Surprisingly, ambiguity of connectives has the lowest correlation coefficient and shows a clear gap with the other factors. This suggests that ambiguity of connectives seems not to be the primary cause of label shift in PDTB 2.0 and PDTB 3.0.

The results above only show the correlation of standalone factors with label shift, without considering all factors together. Inspired by~\citet{liu-etal-2023-whats}, we train an XGBoost model~\citep{xgboost} to find out the importance of factors when using the four features to predict the calculated label shift metric. XGBoost is a gradient boosting framework, where the importance of a feature can be measured by the performance gain it brings~\citep{feaimp}. The framework also harnesses arbitrary interactions between features and is highly regularized to prevent overfitting, making it suitable to analyze a set of features. 

\begin{figure}[t]
\centering\includegraphics[scale=0.385,trim=0 0 0 0]{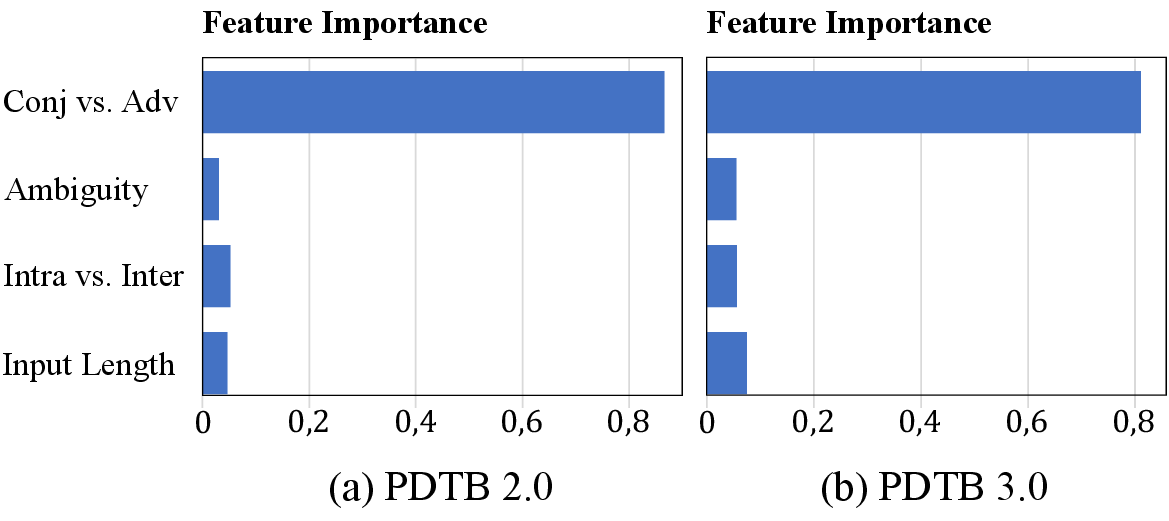}
\setlength{\abovecaptionskip}{-6pt}
\setlength{\belowcaptionskip}{4pt}
\caption{Feature Importance of the XGBoost Model in predicting the label shift metric on PDTB 2.0 and 3.0.}
\label{fig:xgboost}
\end{figure}

We conduct the experiments on PDTB 2.0 and 3.0, and show the results in Figure \ref{fig:xgboost}. Consistent with the Pearson correlation analysis, the syntactic role played by connectives shows overwhelming importance in predicting the label shift metric, with an importance score of more than 0.8. In contrast, the state and length of arguments are less important when all factors are considered together. This may be because the three factors, the syntactic role played by the connective, the state of the arguments, and the length of the arguments, are not independent of each other,\footnote{For example, 62.87\% of explicit examples (in PDTB 2.0) whose connectives are conjunctions, contain intra-sentential arguments. And inter-sentential arguments are usually longer and contain more words than their intra-sentential counterpart.} so the latter two factors provide limited additional information to the first feature in predicting label shift. The last feature, the ambiguity of connective, is still useful but less important than other three factors. We provide more detailed setups and XGBoost results with different combination of features in Appendix \ref{app:xgboost}.

\section{Strategies to alleviate the label shift}
\label{sec:two_strategy}
In this section, we introduce two strategies to alleviate the impact of label shift in the task of explicit to implicit relation recognition. 

\begin{table*}[t]
\centering
\scalebox{0.835}{
\begin{tabular}{l|cccc|cccc}
\hline
\multirow{2}{*}{}          & \multicolumn{4}{c}{PDTB 2.0}\hspace{2.0em}                                     & \multicolumn{4}{|c}{PDTB 3.0}\hspace{2.0em}                                     \\
                           & \multicolumn{2}{c}{top-level}\hspace{1.3em} & \multicolumn{2}{c}{second-level}\hspace{1.3em} & \multicolumn{2}{|c}{top-level}\hspace{1.3em} & \multicolumn{2}{c}{second-level}\hspace{1.3em} \\ \hline
Models                     & Acc\hspace{1.3em}           & F1\hspace{1.3em}            & Acc\hspace{1.3em}             & F1\hspace{1.3em}              & Acc\hspace{1.3em}           & F1\hspace{1.3em}            & Acc\hspace{1.3em}             & F1\hspace{1.3em}             \\ \hline
I2I-Entire                 & 67.97\textsubscript{0.64}         & 59.74\textsubscript{0.94}         & 58.11\textsubscript{0.63}           & 37.74\textsubscript{0.31}           & 72.40\textsubscript{0.21}         & 67.20\textsubscript{0.34}         & 62.62\textsubscript{0.87}           & 53.11\textsubscript{0.58}          \\
I2I-Reduced                   & 63.77\textsubscript{0.53}             & 54.66\textsubscript{1.31}             & 54.07\textsubscript{0.83}               & 35.49\textsubscript{0.49}               & 69.86\textsubscript{0.91}             & 64.12\textsubscript{1.29}             & 59.43\textsubscript{0.40}               & 46.65\textsubscript{0.83}              \\ \hline
\citet{ji-etal-2015-closing}              & -\hspace{1.3em}             & 38.62\hspace{1.3em}         & -\hspace{1.3em}               & -\hspace{1.3em}               & -\hspace{1.3em}             & -\hspace{1.3em}             & -\hspace{1.3em}               & -\hspace{1.3em}              \\
\citet{huang-li-2019-unsupervised}         & -\hspace{1.3em}             & 40.90\hspace{1.3em}         & -\hspace{1.3em}               & -\hspace{1.3em}               & -\hspace{1.3em}             & -\hspace{1.3em}             & -\hspace{1.3em}               & -\hspace{1.3em}              \\
\citet{kurfali-ostling-2021-lets} & -\hspace{1.3em}             & 33.55\hspace{1.3em}         & 25.32\hspace{1.3em}           & 13.01\hspace{1.3em}           & -\hspace{1.3em}             & -\hspace{1.3em}             & -\hspace{1.3em}               & -\hspace{1.3em}              \\ 
Common                     & 53.73\hspace{1.3em}         & 17.48\hspace{1.3em}         & 25.22\hspace{1.3em}           & 03.66\hspace{1.3em}           & 43.62\hspace{1.3em}         & 15.19\hspace{1.3em}         & 27.69\hspace{1.3em}           & 03.10\hspace{1.3em}          \\
E2I-Entire                 & 56.14\textsubscript{0.65}         & 41.49\textsubscript{0.59}         & 34.57\textsubscript{0.38}           & 22.03\textsubscript{0.58}           & 51.49\textsubscript{0.39}         & 45.25\textsubscript{0.50}         & 39.09\textsubscript{0.87}           & 33.56\textsubscript{0.72}          \\
E2I-Reduced                   & 55.58\textsubscript{0.59}             & 39.13\textsubscript{1.05}             & 31.65\textsubscript{0.99}               & 18.03\textsubscript{1.09}               & 48.57\textsubscript{0.30}             & 40.09\textsubscript{0.97}             & 36.54\textsubscript{0.55}            &28.32\textsubscript{1.03}              \\ \hline
Our Method                 & 60.50\textsubscript{0.34}         & 51.25\textsubscript{0.70}         & 39.33\textsubscript{0.28}           & 27.13\textsubscript{0.50}           & 57.54\textsubscript{0.16}         & 51.01\textsubscript{0.45}         & 41.50\textsubscript{0.30}           & 37.08\textsubscript{0.13}          \\ 
 \; w/o filtering                & 58.70\textsubscript{0.24}         & 45.39\textsubscript{0.63}         & 36.28\textsubscript{0.27}           & 23.55\textsubscript{0.53}           & 52.24\textsubscript{0.32}         & 46.03\textsubscript{0.67}         & 40.45\textsubscript{0.33}           & 34.15\textsubscript{0.48}          \\
  \; w/o joint learning                & 57.74\textsubscript{0.45}         & 44.42\textsubscript{0.83}         & 35.23\textsubscript{0.34}           & 22.50\textsubscript{0.48}           & 52.31\textsubscript{0.38}         & 44.46\textsubscript{0.56}         & 40.11\textsubscript{0.28}           & 33.93\textsubscript{0.30}          \\ \hline
\end{tabular}}
\setlength{\abovecaptionskip}{10pt}
\setlength{\belowcaptionskip}{-4pt} 
\caption{Results on PDTB 2.0 and PDTB 3.0 (with standard deviation). E2I-Entire is the typical setting for explicit to implicit discourse relation recognition, serving as the baseline, and I2I-Entire is the upper bound for implicit relation classification. Our two strategies can effectively close the gap between baseline and upper bound.}
\label{table:res-pdtb2-pdtb3}
\end{table*}

\subsection{Filter out noisy examples}
\label{sec:filter_out}
Our first strategy is straightforward: filtering out examples that may have suffered label shift. Given an explicit corpus, we calculate the cosine value of each example following the approach in Section \ref{sec:measure-ls}, and filter out instances with low values. However, rather than configuring a fixed threshold for the entire corpus, we compute different thresholds per relation class. This is because data with different relations suffers from varying degrees of label shift. We group all cosine values based on the relation of each example, and calculate the averaged value for each group. If the cosine value of an instance is less than the average value of the group it belongs to, we filter it out.

\subsection{Joint learning with connectives}
We further investigate a joint learning framework to alleviate label shift in cases where the filtering result is imperfect. The main idea is that label shift is caused by removing connectives; so if we attempt to recover the discarded connective during training, examples may be more consistent with the original relation labels.

Given an explicit example $\rm (Arg1,Conn, Arg2, \\Rel)$, we insert a $\rm \left<mask\right>$ token between two arguments, and train a connective class\-ifier to recover a suitable connective ($\rm conn\_pred$) at the masked position. Simultaneously, we train a relation classifier to predict a relation based on both input arguments and the predicted connective, i.e., $\rm (Arg1, conn\_pred, Arg2)$. With the presence of the predicted connective, we hypothesize that the modified input will be closer to the original relation than the former input containing only two arguments, alleviating the occurrence of label shift. We provide a detailed description and implementation of our method in Appendices \ref{sec:app_method} and \ref{sec:app_implement}, respectively.

\section{Experiments}
\label{sec:exp_res}

We conduct experiments to show that our method not only improve the performance of explicit to implicit relation recognition on PDTB 2.0 and 3.0, but also works well on a corpus annotated with RST relations.

\subsection{Baselines and upper bounds}
We evaluate our method on PDTB 2.0~\citep{prasad-etal-2008-penn} and PDTB 3.0~\citep{webber2019penn}, report the mean performance of 5 runs (with different seeds), and compare our method with existing state-of-the-art models on explicit to implicit relation recognition. In addition, we show the performance of several strong baselines and upper bounds:
\begin{itemize}[leftmargin=*]
    \setlength\itemsep{-0.2em}
    \item \textbf{Common}. Always predict the most common label in the training set.
    \item \textbf{E2I-Entire}. Finetune RoBERTa on the entire training set of explicit examples and test on implicit examples. This is the typical setting for explicit to implicit relation recognition.
    \item \textbf{E2I-Reduced}. Similar to E2I-Entire, but the training set is reduced to have the same size as our filtered corpus.
    \item \textbf{I2I-Entire}. This serves as an upper bound, where RoBERTa is finetuned on the entire training set of the implicit examples.
    \item \textbf{I2I-Reduced}. A variant of I2I-Entire, where the training set contains the same number of examples as our filtered corpus.
\end{itemize}

\subsection{Overall results}
The evaluation results on PDTB 2.0 and PDTB 3.0 are shown in Table \ref{table:res-pdtb2-pdtb3}. Classifiers trained on explicit corpora (E2I) perform much worse on implicit relation recognition than those trained on implicit datasets (I2I). For example, on the top-level relations in PDTB 2.0 and PDTB 3.0,   E2I-Entire lags behind I2I-Entire by 18.25\% and 21.95\% in terms of F1 score, respectively. This is in line with previous findings that classifiers trained on explicit examples perform poorly on real implicit relations~\citep{lin-etal-2009-recognizing}. Our method can substantially enhance the performance of explicit to implicit relation recognition, closing the F1 gap with I2I-Entire from 18.25\% to 8.49\% in PDTB 2.0 and from 21.95\% to 16.19\% in PDTB 3.0. These results highlight the effectiveness of our approach for the task, which in turn suggests that label shift is one cause for poor transfer performance from explicit to implicit relations. Despite achieving impressive results, our model still has a lower performance than the upper bound (i.e., I2I-Entire). We suspect this is because even despite our suggested approach for filtering out training examples potentially affected by label shift, there may still be issues due to the fact that (1) explicit and implicit examples have different syntactic structure~\citep{lin-etal-2009-recognizing}; (2) the label distributions in explicit and implicit corpora are very different (see Figure \ref{fig:tnse-label-shift}). These differences may cause other shifts during the transfer from explicit to implicit relation recognition, which we leave for future work.

We then analyze the effectiveness of each module in our method. Specifically, we perform an ablation study in which we systematically remove the filtering strategy, leaving our model trained with only the joint learning strategy. As shown in Table \ref{table:res-pdtb2-pdtb3}, removing the filtering strategy hurts the performance, with F1 scores for top-level relation recognition dropping by 5.86\% and 4.98\% for PDTB 2.0 and 3.0, respectively. Similarly, we eliminate the joint learning from our approach, giving it the same structure as the baseline but training it on the filtered corpus. Without jointly learning with connectives, the performance of our approach degrades (see "w/o joint learning" in Table \ref{table:res-pdtb2-pdtb3}), similar to the case of the filtering strategy. These results demonstrate that both strategies are crucial for achieving good performance. We also find that using each strategy individually only slightly enhances performance, below the effectiveness of combining them. This suggests that: (1) neither strategy can fully mitigate the impact of label shift; (2) the two strategies are complementary to each other since combining them achieves more improvement. 

Furthermore, we analyze whether our filtering strategy can really improve data quality. To this end, we compare the performance of models trained on the same number of examples but from three different sources: our filtered corpus ("w/o joint learning"), sampling from original explicit corpus (E2I-Reduced), and sampling from implicit corpus (I2I-Reduced). Table \ref{table:res-pdtb2-pdtb3} also shows the results. We find that models trained on our filtered corpus perform better than E2I-Reduced, closing the gap with I2I-Reduced. This indicates that the quality of our filtered corpus is better than the equally sized corpus obtained through random sampling.

\subsection{Results on the GUM dataset}
Our approach is based on the analysis of PDTB corpora. To test the generalizability of our approach, we evaluate it on the GUM dataset~\citep{gum-rst}, which is annotated with RST relations. There are different versions of the GUM dataset, in this work we use the v9 version released by the DISRPT 2023 shared task~\citep{braud-etal-2023-disrpt}. However, the GUM dataset does not have a data split of explicit and implicit relations. To address this issue, we employed a rule to divide explicit and implicit examples: if an instance (1) contains two adjacent text units and (2) contains a connective at the beginning of its second text unit, it is identified as an explicit case; otherwise, it is implicit. We provide a more detailed description of the corpus in Appendix \ref{app:dataset}.

Table \ref{table:res-gum9} shows the results of explicit to implicit relation recognition on the GUM dataset. The classifier trained on explicit examples of the GUM dataset (E2I-Entire) performs poorly in implicit relations, lagging behind its counterpart trained on implicit instances (I2I-Entire) more than 24\% in F1 score. Each of our proposed strategies can slightly improve the performance on this corpus, and combining them achieves the best results, with a 5-point improvement in the F1 score over the E2I-Entire baseline. This demonstrates that our approach may generalize to other discourse data sets. 

\begin{table}[t]
\centering
\scalebox{0.88}{
\begin{tabular}{l|cc}
\hline
Models\hspace{5.6em}             & \hspace{1.4em}Acc\hspace{1.2em}         & \hspace{1.2em}F1\hspace{1.4em}          \\ \hline
I2I-Entire         & \hspace{1.4em}62.08\textsubscript{0.41} & \hspace{1.2em}56.81\textsubscript{0.72} \\
I2I-Reduced        & \hspace{1.4em}52.87\textsubscript{0.67} & \hspace{1.2em}46.51\textsubscript{1.24} \\ \hline
Common             &       35.82      &      07.54       \\
E2I-Entire         & \hspace{1.4em}37.80\textsubscript{0.76} & \hspace{1.2em}32.52\textsubscript{1.34} \\
E2I-Reduced        & \hspace{1.4em}36.26\textsubscript{0.87} & \hspace{1.2em}31.28\textsubscript{1.45} \\ \hline
Our Method         & \hspace{1.4em}41.88\textsubscript{0.63} & \hspace{1.2em}37.56\textsubscript{0.92} \\
\; w/o filtering &     \hspace{1.4em}39.90\textsubscript{0.54}        &    \hspace{1.2em}34.15\textsubscript{0.97}         \\
\; w/o joint learning &     \hspace{1.4em}39.04\textsubscript{0.72}        &    \hspace{1.2em}34.75\textsubscript{1.21}         \\ \hline
\end{tabular}}
\setlength{\abovecaptionskip}{10pt}
\setlength{\belowcaptionskip}{2pt} 
\caption{Results on the RST GUM corpus.}
\label{table:res-gum9}
\end{table}

\section{Conclusion}
We find that one cause for the poor transfer performance from explicit to implicit relations is the occurrence of label shift when deleting connectives from explicit examples. We present both manual and empirical evidence to demonstrate the existence of such shift in the explicit corpus. We design a cosine similarity-based metric to measure label shift in the corpus, filter out noisy data, and investigate a joint learning framework to mitigate label shift. Experiments on PDTB 2.0 and PDTB 3.0 demonstrate that training classifiers on the filtered corpus with our joint learning strategy can significantly enhance the performance of explicit to implicit relation recognition. Furthermore, we show that our approach also works well on the GUM dataset, suggesting its generalizability.

\section{Limitations}
In this study, we conduct experiments solely on corpora annotated with PDTB and RST relations. It would be interesting to explore whether our approach is applicable to corpora annotated with other relations, such as relations in Segmented Discourse Representation Theory~\citep[SDRT,][]{Asher2003-ASHLOC}. In addition, this work only focuses on English relational corpora. Recently, an increasing number of works have begun to call for attention to multilingual discourse~\citep{kurfali-ostling-2019-zero,varachkina-pannach-2021-unified,liu-etal-2023-hits,metheniti-etal-2023-discut}, and shared tasks~\citep{zeldes-etal-2021-disrpt,braud-etal-2023-disrpt} have been organized to deal with multilingual discourse relation classification. Therefore, investigating whether the same findings hold for discourse treebanks in other languages would be an exciting direction for research.

\section*{Acknowledgements}
The authors would like to thank the four anonymous reviewers for their comments. We also thank Xiyan Fu for her valuable feedback on earlier drafts of this paper. This work has been funded by the Klaus Tschira Foundation, Heidelberg, Germany. 

\bibliography{main.bbl}
\bibliographystyle{acl_natbib}

\appendix
\clearpage

\section{Experimental settings}
\label{app:setup}

\subsection{Dataset description}
\label{app:dataset}
\noindent \textbf{PDTB.} The dataset used for analysis in this study is the Penn Discourse Treebank (PDTB). PDTB has two widely used versions, namely PDTB 2.0~\citep{prasad-etal-2008-penn} and PDTB 3.0~\citep{webber2019penn}. In both versions, each example is annotated with a three-level relation from coarse to fine. In this study, we use top-level and second-level relations for analyses and experiments. Following previous work~\citep{kim-etal-2020-implicit}, we use 4 and 11 labels for top- and second-level relations in PDTB 2.0, and 4 and 14 labels for those in PDTB 3.0 (see Table \ref{table:relations}). The two datasets are divided into training, development, and test sets, following the setup in~\citet{ji-eisenstein-2015-one}. Table \ref{table:stat-corpus} shows the statistics information on both datasets. We also show an example of the annotated connective in the implicit corpus:
\vspace{0pt}
\begin{figure}[h]
\centering\includegraphics[scale=0.49,trim=0 0 0 0]{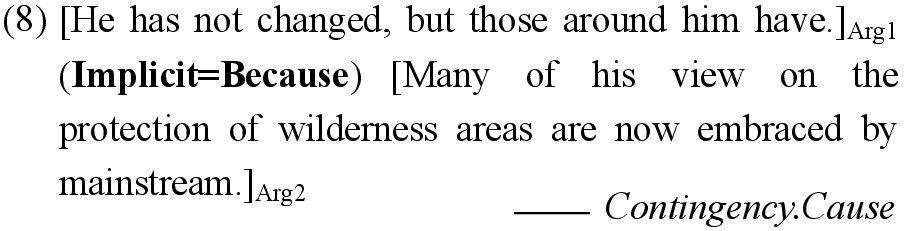}
\vspace{-18pt}
\end{figure}

\begin{table}[t]
\centering
\scalebox{0.72}{
\begin{tabular}{ll} \hline
PDTB 2.0                    & PDTB 3.0        \\ \hline
Comparison                  & Comparison      \\
Contingency                 & Contingency     \\
Expansion                   & Expansion       \\
Temporal                    & Temporal        \\ \hline
Comparison.Concession       & Comparison.Concession  \\
Comparison.Contrast         & Comparison.Contrast    \\
Contingency.Cause           & Contingency.Cause      \\
Contingency.Pragmatic cause & Contingency.Cause+Belief \\
Expansion.Conjunction       & Contingency.Condition  \\
Expansion.Instantiation     & Contingency.Purpose    \\
Expansion.Alternative       & Expansion.Conjunction  \\
Expansion.List              & Expansion.Equivalence  \\
Expansion.Restatement       & Expansion.Instantiation  \\
Temporal.Asynchronous       & Expansion.Level-of-detail \\
Temporal.Synchrony          & Expansion.Manner       \\
            & Expansion.Substitution    \\
            & Temporal.Asynchronous     \\
            & Temporal.Synchronous   \\  \hline
\end{tabular}}
\setlength{\abovecaptionskip}{10pt}
\setlength{\belowcaptionskip}{2pt}
\caption{Top- and second-level relations of PDTB 2.0 and PDTB 3.0 (commonly used in the literature).}
\label{table:relations}
\end{table}

\begin{table}[t]
\centering
\scalebox{0.72}{
\begin{tabular}{llll}
\hline
Joint     & Adversative   & Context  & Causal\hspace{1.2em} \\
Elaboration\hspace{0.9em}   & Explanation\hspace{0.9em} & Contingency\hspace{0.9em} & \\ \hline
\end{tabular}}
\setlength{\abovecaptionskip}{10pt}
\setlength{\belowcaptionskip}{2pt}
\caption{Relations of the GUM dataset used in this work.}
\label{table:gum_rel}
\end{table}

\noindent \textbf{GUM}. The GUM dataset used in this study is from DISRPT 2023~\citep{braud-etal-2023-disrpt}. The original GUM dataset is annotated with a constituent tree structure, containing both structure and relation information. In DISRPT 2023, the annotated relations in the GUM v9 dataset have been converted into a group of triplets which contain text units 1 and 2, and label. Specifically, the organizers of DISRPT 2023 first convert\footnote{\url{https://github.com/amir-zeldes/gum}} the RST constituent trees into a dependency representation~\citep{li-etal-2014-text}. They then extract (EDU\textsubscript{i}, EDU\textsubscript{i}, Relation) from the dependency tree as (text\_unit1, text\_unit2, label)\footnote{We consider the text unit as argument.} in the shared task. However, there is no split between explicit and implicit relations in this corpus. To address this issue, we follow previous research practice on RST corpus~\citep{marcu-echihabi-2002-unsupervised} to consider examples that have two adjacent text units and contain a connective at the begining of the second text unit as explicit instances. The connective list used during the division comes from the explicit corpus of PDTB 2.0, which includes 99 distinct connectives. The processed corpus contains about 3k explicit examples and 19.2k implicit instances. Due to the small size of the explicit corpus and the uneven distribution of labels (e.g., the relation \textit{joint} accounts for 34.8\% of the explicit corpus), we only consider relations with frequency more than 100, resulting in 7 relations (see Table \ref{table:gum_rel}) in the final dataset. We show the statistics of the final dataset in Table \ref{table:stat-corpus}.

\begin{table}[t]
\centering
\scalebox{0.8}{
\begin{tabular}{l|c|ccc}
\hline
Dataset                   & \hspace{0.9em}Type\hspace{0.9em}     & \hspace{0.4em}Train\hspace{0.4em} & \hspace{0.4em}Dev\hspace{0.4em} & \hspace{0.4em}Test\hspace{0.4em} \\ \hline
\multirow{2}{*}{PDTB 2.0} & Explicit &   14117    &  1462   &   1285   \\
                          & Implicit &   12632    &  1183   &   1046   \\ \hline
\multirow{2}{*}{PDTB 3.0} & Explicit &   18626    &  1944   &   1767   \\
                          & Implicit &   17085    &  1653   &   1474   \\ \hline
\multirow{2}{*}{GUM}      & Explicit &   2095    &  276   &  264    \\
                          & Implicit &   11802    &  1607   &   1619   \\ \hline
\end{tabular}}
\setlength{\abovecaptionskip}{8pt}
\setlength{\belowcaptionskip}{0pt} 
\caption{Statistics of different corpus.}
\label{table:stat-corpus}
\end{table}

\begin{figure}[t]
\centering\includegraphics[scale=0.29,trim=0 0 0 0]{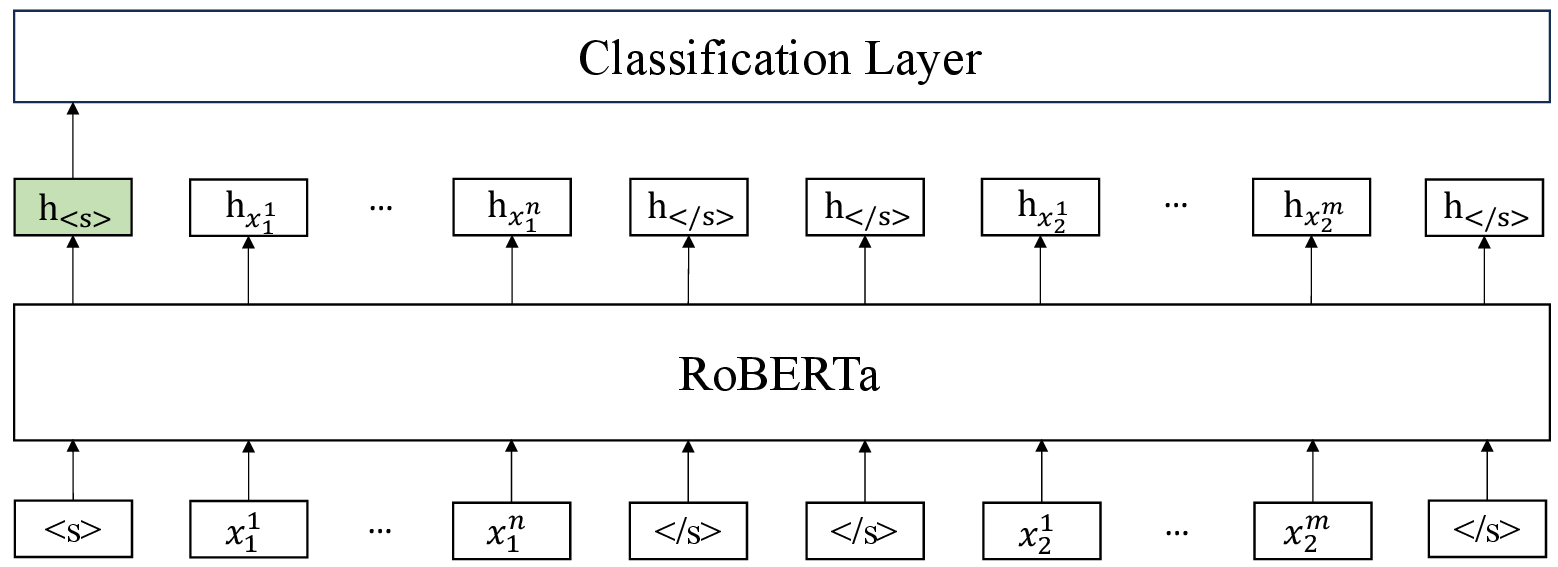}
\setlength{\abovecaptionskip}{0pt}
\setlength{\belowcaptionskip}{10pt}
\caption{The architecture of the model used in the analysis.}
\label{fig:model_base}
\vspace{-10pt}
\end{figure}

\begin{table*}[t]
\centering
\scalebox{0.85}{
\begin{tabular}{l|l|c|cc|cc}
\hline
\multirow{2}{*}{Encoder Type}      & \multirow{2}{*}{Relation Level} & \multirow{2}{*}{Relation Type} & \multicolumn{2}{c|}{PDTB 2.0} & \multicolumn{2}{c}{PDTB 3.0} \\ \cline{4-7} 
                                   &                                 &                                & \hspace{1em}Same\hspace{1em}        & Different       & \hspace{1em}Same\hspace{1em}        & Different      \\ \hline
\multirow{2}{*}{RoBERTa-base}      & \multirow{2}{*}{Second-level}   & Explicit                       & 59.21       & 40.79           & 67.56       & 32.44          \\
                                   &                                 & Implicit                       & 93.67       & 6.33            & 93.14       & 6.86           \\ \hline
\multirow{2}{*}{RoBERTa-large}     & \multirow{2}{*}{Top-level}      & Explicit                       & 59.81       & 40.19           & 66.84       & 33.16          \\
                                   &                                 & Implicit                       & 93.73       & 6.27            & 94.27       & 5.73           \\ \hline
\multirow{2}{*}{BERT-base-uncased} & \multirow{2}{*}{Top-level}      & Explicit                       & 59.44       & 40.56           & 70.52       & 29.48          \\
                                   &                                 & Implicit                       & 94.53       & 5.47            & 95.97       & 4.03           \\ \hline
\multirow{2}{*}{DeBERTa-base} & \multirow{2}{*}{Top-level}      & Explicit                       & 64.64       & 35.36           & 69.41       & 30.59          \\
                                   &                                 & Implicit                       & 96.08       & 3.92            & 94.88       & 5.12           \\ \hline
\multirow{2}{*}{T5-base} & \multirow{2}{*}{Top-level}      & Explicit                       & 63.41       & 36.59           & 69.86       & 30.14          \\ 
                                   &                                 & Implicit                       & 94.68       & 5.32            & 94.49       & 5.51           \\
                                    \hline
\end{tabular}}
\setlength{\abovecaptionskip}{12pt}
\setlength{\belowcaptionskip}{-6pt}
\caption{Percentage of examples in Explicit and Implicit corpora that receive the same and different predictions when containing and not containing a connective (see $\rm diff\_num$ in Algorithm \ref{al:measuring-ls}).}
\label{tab:ls-level-type-size}
\end{table*}

\subsection{Details in analyses}
\label{app:ana-detail}
The model used in the analysis is a widely used framework for relation classification, consisting of a pre-trained model as the encoder and a linear layer as the classification layer. We follow previous work~\citep{zhou-etal-2022-prompt-based,long-webber-2022-facilitating} to use $\rm RoBERTa_{base}$ as the encoder. Figure \ref{fig:model_base} shows the overall architecture of the model. We train this model following most of the default settings in RoBERTa. The optimizer used in the experiments is AdamW, with an initial learning rate of 1e-5, a batch size of 16, and a maximum epoch number of 10. The maximum input length is set to 256 for all corpora. We conduct all experiments on a single Tesla P40 GPU with 24GB memory. Note that the baselines in the experimental section (e.g., E2I-Entire, E2I-Reduced, I2I-Entire, and I2I-Reduced) use the same models as described above.

\section{More results about label shift}
\subsection{Manual analysis}
\label{app:manu_ana}
We randomly sample 100 explicit examples\footnote{Among the 100 examples, 16, 31, 34, and 19 are in Contingency, Comparison, Expansion, and Temporal Relations, respectively, similar to the label distribution of the explicit corpus (Contingency: 17.74\%, Comparison: 29.82\%, Expansion: 33.79\%, Temporal: 18.65\%).} from PDTB 2.0 and remove the connective from those examples. Before starting the annotation, two students from the Computational Linguistics Department receive training in annotating relations on unannotated texts. Specifically, we introduce the definition of relations in PDTB 2.0 to the two students and ask them to practice annotating raw texts from Gigaword corpus (the corpus, like PDTB 2.0, is also in news domain). We check their annotation results, compare the difference, listen to their explanations, and give our comments. After several rounds of practice, the two students are asked to annotate these 100 explicit examples without connectives, separately. They need to annotate each instance with one of 12 relations, including \textit{Comparison.Concession}, \textit{Comparison.Contrast}, \textit{Contingency.Cause}, \textit{Contingency.Pragmatic cause}, \textit{Expansion.Conjunction}, \textit{Expansion.Instantiation}, \textit{Expansion.Alternative}, \textit{Expansion.List}, \textit{Expansion.Restatement}, \textit{Temporal.Asynchronous}, \textit{Temporal.Synchrony}, and \textit{NonRel}. The inter-annotator agreement is 0.7346 calculated in Cohen's kappa.

\setlength{\textfloatsep}{12pt}
\begin{algorithm}[t]
    \small
    \renewcommand{\algorithmicrequire}{\textbf{Input:}}
    \renewcommand{\algorithmicensure}{\textbf{Output:}}
    \caption{Identify Cases of Label Shift}
    \begin{algorithmic}[1]
        \Require
            New and original annotations $\rm \{(NA_i, OA_i)\}|_{i=1}^{100}$  
        \Ensure
            Statistics of label shift: $\rm shift\_num$, $\rm case1$, $\rm case2$, $\rm case3$

            \State $\rm shift\_num, case1, case2, case3=0, 0, 0, 0$ 
            \For{$\rm i=1, \dots, 100$}
                \If{$\rm NA_i \neq OA_i$}
                    \State $\rm shift\_num \mathrel{+}= 1$
                    \State 
                    \If{$\rm OA_i \in NA_i$}
                        \State $\rm case2 \mathrel{+}= 1$
                    \ElsIf{$\rm NA_i=$"NoRel"}
                        \State $\rm case3 \mathrel{+}= 1$
                    \Else{}
                        \State $\rm case1 \mathrel{+}= 1$
                    \EndIf
                \EndIf
            \EndFor
    \end{algorithmic}  
    \label{al:shift_case}
\end{algorithm}

After completing the annotations, we implement a program (see Algorithm \ref{al:shift_case}) following the definition of label shifting in equation (\ref{eqt:shift}), comparing the new annotations with the original ones\footnote{The original labels of these 100 explicit examples are annotated based on both argument and connective. And the inter-annotator agreement of the original explicit corpus is 0.945 ~\citep{prasad-etal-2008-penn}.}. If an example's new and original annotations are different (not exactly equal), it is considered to suffer a label shift. Further, given an example suffering a label shift, it is case (ii) if its new annotation contains the original one; otherwise, it is case (iii) if the new annotation is "no relation"; otherwise, it is case (i). Among the 100 examples, 37 are identified as suffering a label shift, in which 15, 19, and 3 belong to types (i), (ii), and (iii), respectively. 

\subsection{Empirical evidence}
\label{app:diff_type_size}
Most of the analytical results in Section \ref{sec:ls_corpus} are based on $\rm RoBERTa_{base}$ and only cover top-level relations. Here, we first show the analytical results on second-level relations with $\rm RoBERTa_{base}$. Then, we present the top-level results based on a larger size of pre-trained models, such as $\rm RoBERTa_{large}$, or with other types of pre-trained models, such as $\rm BERT$, $\rm DeBERTa$, and $\rm T5$. We aim to demonstrate that consistent conclusions can be drawn regardless of relation level, pre-trained model type, or scale.

Table \ref{tab:ls-level-type-size} shows the results on PDTB 2.0 and PDTB 3.0. We observe similar results under different settings to Figure \ref{fig:sen_non-sen}. That is, the explicit corpus has more examples receiving different predictions when containing and not containing a connective. 

\subsection{More XGBoost results}
\label{app:xgboost}
\noindent \textbf{Setting}. Given an explicit corpus with annotated relations $\rm \{(Arg1_i, Conn_i, Arg2_i, Rel_i)\}|_{i=1}^{N}$, we first calculate the label shift metric for each example as shown in Section \ref{sec:measure-ls}. We then extract four features from each explicit example:
\begin{itemize}
    \item[1. ] Is the included connective is a conjunction or an adverb? 1 for conjunction and 0 for adverb. 
    \item[2. ] Is the included connective ambiguous? 1 for ambiguous\footnote{We consider connectives that can signal more than one discourse relation as ambiguous. See Appendix A in~\citet{Prasad2006ThePD} and~\citet{webber2019penn}.} and 0 for not ambiguous. 
    \item[3.] Is the status of the contained arguments intra- or inter-sentential? 1 for intra-sentential and 0 for inter-sentential. 
    \item[4.] The number of words included in the argumen\-ts. We normalize the value between 0 and 1.
\end{itemize}

\begin{table}[t]
\centering
\scalebox{0.78}{
\begin{tabular}{|lc|lc|}
\hline
\multicolumn{2}{|c|}{Two Featues} & \multicolumn{2}{c|}{Three featues} \\ \hline
Feature            & Importance   & Feature            & Importance    \\ \hline
Conj vs.  Adv      & 0.9950       & Conj vs.  Adv      & 0.9807        \\
Ambiguity          & 0.0050       & Ambiguity          & 0.0036        \\ \cline{1-2}
Conj vs.  Adv      & 0.9834       & Intra vs. Inter    & 0.0156        \\ \cline{3-4} 
Intra vs. Inter    & 0.0166       & Conj vs.  Adv      & 0.9481        \\ \cline{1-2}
Conj vs.  Adv      & 0.9717       & Ambiguity          & 0.0159        \\
Input length       & 0.0283       & Input length       & 0.0360        \\ \hline
Ambiguity          & 0.1456       & Conj vs.  Adv      & 0.9212        \\
Intra vs. Inter    & 0.8544       & Intra vs. Inter    & 0.0374        \\ \cline{1-2}
Ambiguity          & 0.4543       & Input length       & 0.0414        \\ \cline{3-4} 
Input length       & 0.5457       & Ambiguity          & 0.1853        \\ \cline{1-2}
Intra vs. Inter    & 0.8925       & Intra vs. Inter    & 0.7058        \\
Input length       & 0.1075       & Input length       & 0.1089        \\ \hline
\end{tabular}}
\setlength{\abovecaptionskip}{12pt}
\setlength{\belowcaptionskip}{-2pt}
\caption{Feature importance from XGBoost when inputting different combinations of features.}
\label{tab:fea_comb}
\end{table}

We first calculate the Pearson correlation~\cite{liu-etal-2023-modeling} between each individual feature and the label shift metric at the corpus level. We then train an XGBoost model using features to predict the label shift metric and analyze the importance of each feature. Table \ref{tab:fea_comb} shows the results when inputting different combinations of features into XGBoost. The syntactic role played by connectives contributes the most in predicting the occurrence of the label shift. When the feature of the syntactic role of connectives is not considered, the status of the arguments becomes the most important feature in predicting the label shift metric. The length of arguments and the ambiguity of connectives contribute similarly to label shift but are less important than the syntactic role played by connectives and the status of arguments.

Our analysis results reveal that the syntactic role played by the removed connective is an important factor in the occurrence of label shift. In PDTB, connectives come from two grammatical categories: Conjunctions (including subordinating conjunctions and coordinating conjunctions) and Adverbs~\citep{Prasad2006ThePD}. Conjunctions are used to grammatically connect clauses~\citep{gram-conjunction}, removing them can render the entire text ungrammatical and unclear in expression. For example, if we remove the \textit{and} in "[Mr. Stein moved to a new city] and [he found a job there]", the text becomes ungrammatical and we can not know whether the text wants to express a \textit{Conjunction} relation (with a \textit{and}) or a \textit{Cause} relation (with a \textit{because}). By contrast, adverbs typically link two sentences, aiming to facilitate communication rather than serving grammatical purposes~\citep{gram-adverbial}. The elimination of adverbs may lead to reduced coherence in explicit examples, but its meaning is generally unchanged. For instance, if we remove the \textit{however,} from "[Such problems will require considerable skill to resolve.] However, [neither Mr. Baum nor Mr. Harper has much international experience.]", the entire text becomes less coherent, but the expressed relation remains \textit{Contrast}. 

\section{Strategies to alleviate the label shift}
\label{sec:app_method}
\subsection{Filter out noisy examples}
Given an explicit relation corpus $\rm \{(Arg1_i, Conn_i, \\ Arg2_i, Rel_i)\}|_{i=1}^{N}$, we calculate the cosine similarity metric (see $\rm scores$ in Algorithm \ref{al:measuring-ls}) for each example and obtain scores $\rm \{s_1, ..., s_N\}$. We then calculate the average scores grouped by different relations. If the cosine value of an instance is less than the average value of the group it belongs to, we filter it out (see Algorithm \ref{al:filtering}). 

\setlength{\textfloatsep}{12pt}
\begin{algorithm}[t]
    \small
    \renewcommand{\algorithmicrequire}{\textbf{Input:}}
    \renewcommand{\algorithmicensure}{\textbf{Output:}}
    \caption{Filtering Sensitive Examples}
    \begin{algorithmic}[1]
        \Require
            Examples with scores $\rm \{(E_i, Rel_i, s_i)\}|_{i=1}^{N}$  
        \Ensure
            Filtered corpus $\rm C$

            \State $\rm groups=\{\}$ 
            \State $\rm threshold = \{\}$ 
            \State $\rm C=[]$ 
            \For{$\rm i=1, \dots, N$}
                \If{$\rm Rel_i\;in\;groups $}
                    \State $\rm Append(groups[Rel_i], s_i)$
                \Else
                    \State $\rm groups[Rel_i]=[s_i]$
                \EndIf
            \EndFor
            \State
            \For{$\rm Rel \;in\;groups$}
                \State $\rm threshold[Rel]=Avg(groups[Rel])$
            \EndFor
            \State 
            \For{$\rm i=1, \dots, N$}
                \If{$\rm s_i \ge threshold[Rel_i] $}
                    \State $\rm Append(C, E_i)$
                \EndIf
            \EndFor
    \end{algorithmic}  
    \label{al:filtering}
\end{algorithm}

\subsection{Joint learning with connectives}
\begin{figure}[t]
\centering
\includegraphics[scale=0.305,trim=0 0 0 0]{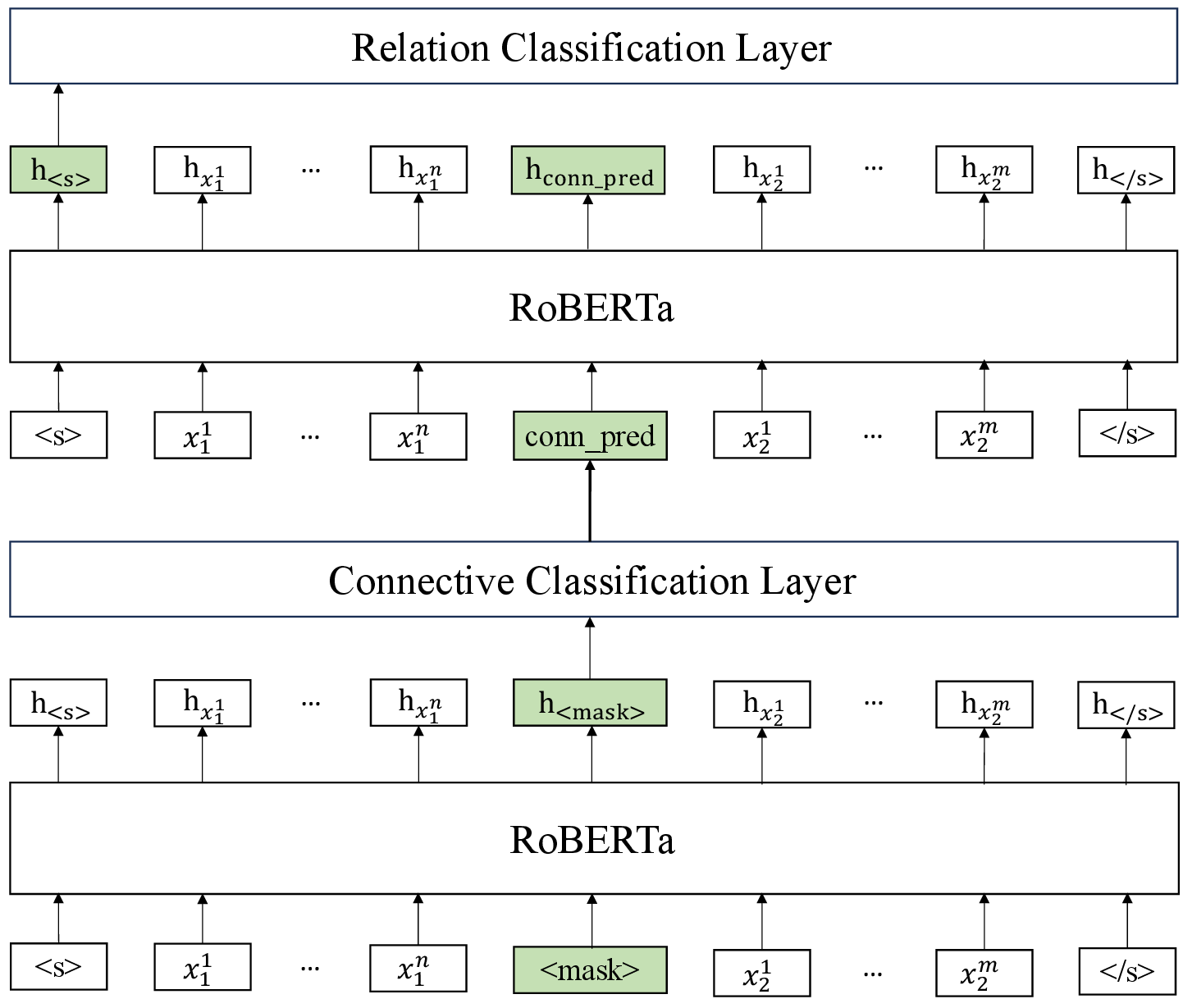}
\setlength{\abovecaptionskip}{12pt}
\setlength{\belowcaptionskip}{0pt}
\caption{The architecture of the joint learning model.}
\label{fig:model_joint}
\end{figure}

Inspired by recent work using connective information for relation classification~\citep{kishimoto-etal-2020-adapting,zhou-etal-2022-prompt-based,liu-strube-2023-annotation}, we investigate a joint learning framework for explicit to implicit relation recognition. Specifically, we jointly train the model to recover a connective between arguments and to predict a relation based on the recovered connective\footnote{We can not directly input the golden connectives to the relation classifier for training. Because this will make the classifier rely heavily on golden connective for prediction~\citep{pitler-nenkova-2009-using}, which results in poor evaluation performance on implicit corpus. In the task of explicit to implicit relation recognition,  golden connectives are not available during evaluation. } and arguments. Figure \ref{fig:model_joint} shows the overall architecture of the joint learning model.

Given an explicit example $\rm (Arg1, Conn, Arg2,\\ Rel)$, where $\rm Arg1=\{x_1^1,...,x_1^{n}\}$, $\rm Arg2=\{x_2^1, ...\\, x_2^m\}$, we insert an $\rm \left< mask \right>$ token between arguments, input them to $\rm RoBERTa$, and train the mod\-el to predict a connective at the masked position:
\begin{equation}
    \mathbf{p}^c = {\rm softmax(\mathbf{W}_c h_{\rm \left< mask \right>}+\mathbf{b}_c)}
\end{equation}
where $\rm \mathbf{p}^c$ denotes the probabilities of all connectives. We train this module with cross-entropy loss:
\begin{equation}
    \mathcal{L}_{Conn} = -\sum_{i=0}^{N} \sum_{j=0}^{CN} C_{ij} \log (P_{ij}^c)
\end{equation}
where $C_{i}$ is the ground-truth connective of the $i$-th instance with a one-hot scheme, $CN$ is the total size of connectives. 

Simultaneously, we train the model to predict a relation based on both the predicted connective and arguments. To achieve so, we adopt the Gumbel-Softmax to sample a connective $\rm conn\_pred$ at the masked position:
\begin{equation}
    \label{equ:gumbel}
    \begin{aligned}
        g &= -\log(-\log(\xi)), \; \xi \sim {\rm U}(0,1) \\
        \mathbf{c}_i &= \frac{\exp((\log(\textbf{p}_i^c)+g_i)/\tau)}{\sum_j \exp((\log(\textbf{p}_j^c)+g_j)/\tau)}
    \end{aligned}
\end{equation}
where $g$ is the Gumbel distribution, $\rm U$ is the uniform distribution, $\mathbf{p}_i^c$ is the probability of $i$-th connective output by the connective classifier, $\tau \in (0, \infty)$ is a temperature parameter (we set $\tau = 1.0$ in experiments). We use the Gumbel-Softmax rather than a normal $\rm arg max$ operation because the former enables the end-to-end training of the whole model, alleviating cascading errors caused by incorrectly predicted connectives. Next, we replace the $\rm \left< mask \right>$ token with the predicted connective $\rm conn\_pred$, feed them to $\rm RoBERTa$, predict a relation using the hidden states of the first token:
\begin{equation}
    \mathbf{p}^r = {\rm softmax}(\mathbf{W}_r h_{\rm \left< s \right>}+\mathbf{b}_r)
\end{equation}
and train this module with cross-entropy loss:
\begin{equation}
    \setlength{\abovedisplayskip}{6pt}
    \setlength{\belowdisplayskip}{6pt}
    \mathcal{L}_{Rel} = -\sum_{i=0}^{N} \sum_{j=0}^{RN} Y_{ij} \log (P_{ij}^r)
\end{equation}
where $Y_{i}$ is the ground-truth relation of the $i$-th example with a one-hot scheme, $RN$ is the total size of relations.

We jointly train the two modules with a multi-task loss:
\begin{equation}
    \mathcal{L} = 0.5 * \mathcal{L}_{Conn} + \mathcal{L}_{Rel}
\end{equation}
Since our primary goal is relation prediction, we give a larger weight to relation loss $\mathcal{L}_{Rel}$.

\section{Implementation details}
\label{sec:app_implement}

We have introduced the details of E2I and I2I baselines in Appendix \ref{app:ana-detail}, here we mainly focus on the implementation of our approach.

For data filtering, we use the averaged cosine similarity score of each relation group as the threshold. This works well for PDTB 2.0 and PDTB 3.0 but we made a small adjustment to the settings for the GUM corpus. Specifically, we filter out an instance (in the GUM corpus) only if its cosine similarity score is lower than the average value of the group it belongs to and its cosine similarity score is less than 0.6. We do so because the size of the GUM corpus is small (about 2k, see Table \ref{table:stat-corpus}). If we filter out too many instances, there will not be enough data to train classifiers to converge.

Regarding the joint learning, we use barely the same settings as baselines, including $\rm RoBERTa_{base}$, AdamW optimizer, batch size of 16, learning rate of 1e-5, a maximum training epoch of 10, and maximum input length of 256.

\section{Discussion of threshold}
In Section \ref{sec:filter_out}, we propose to use an averaging strategy to compute thresholds for different relations (called relation average). It is motivated by two observations. (1) There is a trade-off between the size and quality of the filtered corpus. If we use a large threshold, most of the noisy examples will be filtered out, but it may also filter out good examples. As a result, the filtered corpus will be relatively small, affecting the performance of the trained model (i.e., the size of the training set can affect the performance). For example, when we use 0.8 as the threshold for PDTB 2.0 (top-level relations), about 49.89\% examples will be filtered out, and our full model trained on this filtered corpus can only achieve an F1-score of 46.24. If we use a small threshold, only a few noisy examples will be filtered out, but the size of the filtered corpus is close to the original corpus. In extreme cases, if we do not filter out any examples, our model will degrade into the 'w/o filtering' ablation version in Table \ref{table:res-pdtb2-pdtb3}. Therefore, we propose to use the average strategy to achieve a rough balance between quality and size. (2) Data with different relations suffers from varying degrees of label shift (see Fig \ref{fig:tnse-label-shift}a and \ref{fig:tnse-label-shift}b). We also tried another type of average strategy (called global average): calculate the average cosine value of all examples and use it as the threshold. On the PDTB 2.0 top-level relations, the F1 score using the global average is 48.12, lower than 51.23 using the relation average. Therefore, we chose the relation average in our paper. Tables \ref{table:res-pdtb2-pdtb3} and \ref{table:res-gum9} show it works well on those corpora. Since the primary goal of Sections \ref{sec:two_strategy} and \ref{sec:exp_res} is to demonstrate that our finding of label shifting is also helpful for improving performance, we leave the exploration of better threshold selection for future work.

\end{document}